\documentclass[twoside,11pt]{article} 
\usepackage[hyphens]{url}
\usepackage[implicit=true,unicode=true,
 bookmarks=true,bookmarksnumbered=true,bookmarksopen=true]
 {hyperref}
\usepackage{jair, theapa, rawfonts}
\usepackage{graphicx}
\usepackage{booktabs}
\usepackage{multirow}
\usepackage{array}
\usepackage{varwidth}
\usepackage[table]{xcolor}
\usepackage[table]{xcolor}

\definecolor{oran}{RGB}{245, 210, 153}

\usepackage{enumitem}
    \setlist[itemize]{wide=0pt}

\newcolumntype{C}{>{\centering\arraybackslash}m{38mm}}

\newcommand{\fake}[1]{$\textrm{#1}$}

\begin{document}

\title{Reinforcement Learning for Generative AI: State of the Art, Opportunities and Open Research Challenges}

\author{\name Giorgio Franceschelli \email giorgio.franceschelli@unibo.it \\
       \addr Department of Computer Science and Engineering, University of Bologna, Bologna, Italy
       \AND
       \name Mirco Musolesi \email m.musolesi@ucl.ac.uk \\
       \addr Department of Computer Science, University College London, London, United Kingdom\\
       \addr Department of Computer Science and Engineering, University of Bologna, Bologna, Italy}


\maketitle

\begin{abstract}
Generative Artificial Intelligence (AI) is one of the most exciting developments in Computer Science of the last decade. At the same time, Reinforcement Learning (RL) has emerged as a very successful paradigm for a variety of machine learning tasks. In this survey, we discuss the state of the art, opportunities and open research questions in applying RL to generative AI. In particular, we will discuss three types of applications, namely, RL as an alternative way for generation without specified objectives; as a way for generating outputs while concurrently maximizing an objective function; and, finally, as a way of embedding desired characteristics, which cannot be easily captured by means of an objective function, into the generative process. We conclude the survey with an in-depth discussion of the opportunities and challenges in this fascinating emerging area.
\end{abstract}

\section{Introduction}
\label{Introduction}

Generative Artificial Intelligence (AI) is gaining increasing attention in academia, industry, and among the general public. 
This has been apparent since a portrait based on Generative Adversarial Networks \shortcite{goodfellow14} was sold for more than four hundred thousand dollars\footnote{\url{www.christies.com/features/a-collaboration-between-two-artists-one-human-one-a-machine-9332-1.aspx}} in 2018. Then, the introduction of transformers \shortcite{vaswani17} for natural language processing and diffusion models \shortcite{sohldickstein15} for image generation has led to the development of generative models characterized by unprecedented performance, e.g., GPT-4 \shortcite{openai23}, LaMDA \shortcite{thoppilan22}, Llama 2 \shortcite{touvron2023}, Gemini \shortcite{gemini23}, DALL-E 2 \shortcite{ramesh22} and Stable Diffusion \shortcite{rombach22}, just to name a few. In particular, ChatGPT\footnote{\url{https://openai.com/blog/chatgpt/}}, a conversational agent based on GPT-3 and GPT-4, is widely considered as a game-changing product; its introduction has indeed accelerated the development of foundation models. 
One of the characteristics of ChatGPT and other state-of-the-art large language models (LLMs) and foundation models\footnote{We assume the following definitions: we refer to large language models as language models characterized by large size in terms of number of parameters; they are are also usually based on transformer architectures. A foundation model is a large model that is trained on broad data of different types (textual, audio, image, video, etc.) at scale and is adaptable to a wide range of downstream tasks, following \shortciteA{bommasani22}.} is the use of Reinforcement Learning (RL) in order to align its production to human values \shortcite{christiano17}, so as to mitigate biases and to avoid mistakes and potentially malicious uses. 

In general, RL offers the opportunity to use non-differentiable functions as rewards \shortcite{ranzato16}. Examples include chemistry \shortcite{vanhaelen20} and dialogue systems \shortcite{young13}. We believe that RL is a promising solution for designing efficient and effective generative AI systems. In this article, we will explore this research space, which is, after all, largely unexplored.
In particular, the contributions of this work can be summarized as follows: we first survey the current state of the art at the interface (and intersection) between generative AI and RL. We systematize the existing literature according to three classes of applications, namely RL as an alternative way for generation with the goal of approximating outputs in the domain of interest as best as possible; as a way for generating outputs while concurrently maximizing quantifiable metrics or indicators; and, finally, as a way of embedding desired characteristics, which cannot be easily captured by means of an objective function, into the generative process. We then discuss the future opportunities and challenges of each category, outlining a potential research agenda for the coming years.

Several works have already surveyed deep generative learning (e.g., \shortciteR{franceschelli21,foster23}), deep reinforcement learning (e.g., \shortciteR{lazaridis20,sutton18}), its societal impacts \shortcite{whittlestone21}, and applications of RL for specific generative domains (e.g., \shortciteR{fernandes23}). To the best of our knowledge, this is the first survey on the applications (and implications) of RL applied to generative deep learning.

The remainder of the paper is structured as follows. First, we introduce and review key concepts in generative AI and RL (Section \ref{preliminaries}). Then, we discuss the different ways in which RL can be used for generative tasks, both considering past works and suggesting future directions (Section \ref{GRL}). Finally, we conclude the survey by discussing open research questions and analyzing future research opportunities (Section \ref{conclusion}).

\section{Preliminaries} \label{preliminaries}

\subsection{Generative Deep Learning} \label{gdl}

We will assume the following definition of \textit{generative model} \shortcite{foster23}: given a dataset of observations $X$, and assuming that $X$ has been generated according to an unknown distribution $P_{data}$, a generative model $P_{model}$ is a model that can mimic $P_{data}$. By sampling from $P_{model}$, observations that appear to have been drawn from $P_{data}$ can be generated. Generative deep learning consists in the application of deep learning techniques to learn $P_{model}$.

Several families of generative deep learning techniques have been proposed in the last decade, e.g., Variational Autoencoders \shortcite<VAEs;>{kingma14,rezende14}, Generative Adversarial Networks \shortcite<GANs;>{goodfellow14}, autoregressive models like Recurrent Neural Networks \shortcite<RNNs;>{cho14,hochreiter97}, transformers \shortcite{vaswani17}, and denoising diffusion models \shortcite{sohldickstein15,ho20}. These models and architectures aim to approximate $P_{data}$ by means of self-supervised learning, i.e., by minimizing a reconstruction error when trying to reproduce real examples from $X$. The only exceptions are GANs, which aim to approximate $P_{data}$ using adversarial learning, i.e., by maximizing the predicted probability that the outputs were generated by $P_{data}$. We refer the interested reader to \shortciteA{franceschelli21} for a deeper analysis of the training and sampling processes at the basis of these solutions. Although highly effective for a variety of tasks, the outputs generated by these models do not always satisfy the desired properties. This happens for a variety of reasons. In fact, specific objectives cannot always be cast as loss functions; and providing carefully designed datasets is typically expensive. Few-shot learning \shortcite{brown20}, prompt engineering \shortcite{strobelt22} and fine-tuning \shortcite{dodge20} are potential solutions to these problems. We will discuss these issues in detail in the following sections.

\subsection{Deep Reinforcement Learning}

\begin{figure}[ht]
  \centering
  \includegraphics[width=0.5\linewidth]{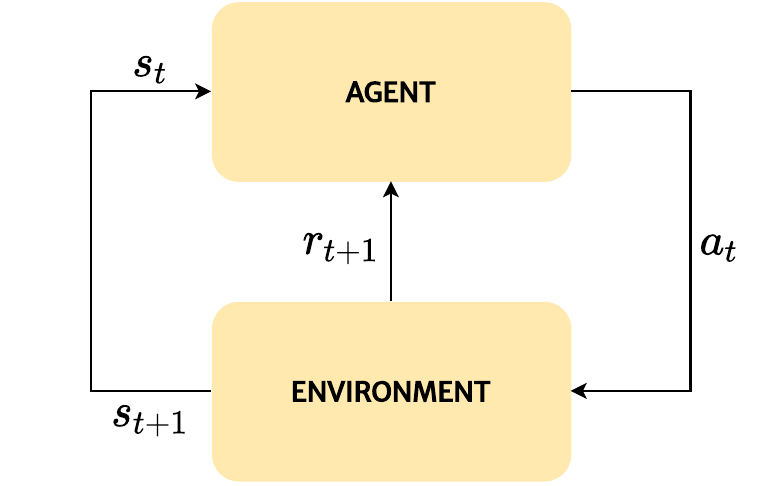}
  \caption[RL Framework]{The canonical reinforcement learning framework: at each timestep $t$, the Agent performs an action $a_t$ based on the current state $s_t$, which is a representation of the Environment. Upon the execution of the action, the Agent finds itself in a new state $s_{t+1}$, and receives a reward $r_{t+1}$.}
  \label{fig:rl}
\end{figure}

RL is a machine learning paradigm that consists in learning an action based on a current representation of the environment in order to maximize a numerical signal, i.e., the \textit{reward} over time \shortcite{sutton18}. More formally, at each time step $t$, an \textit{agent} receives the current \textit{state} from the \textit{environment}, then it performs an \textit{action} and observes the reward and the new state. Figure \ref{fig:rl} summarizes the process. The learning process aims to teach the agent to act in order to maximize the \textit{cumulative return}, i.e., a discounted sum of future rewards. Deep learning is also used to learn and approximate a \textit{policy}, i.e., the mapping from states to action probabilities, or a \textit{value function}, i.e., the mapping from states (or state-action pairs) to expected cumulative rewards. In this case, we refer to it as deep reinforcement learning. Several algorithms have been proposed to learn a value function from which it is possible to induce a policy, e.g., DQN \shortcite{mnih13} and its variants \shortcite{vanhasselt16,schaul16,wang16}, or to directly learn a policy, e.g., A3C \shortcite{mnih16}, DDPG \shortcite{lillicrap16}, TRPO \shortcite{schulman15}, PPO \shortcite{schulman17}. We refer the interested readers to \shortciteA{sutton18} for a comprehensive introduction to the topic.

The RL community has developed a variety of solutions to address the specific theoretical and practical problems emerging from this simple formulation.
For example, if the reward signal is not known, inverse reinforcement learning \shortcite<IRL;>{ng00} is used to learn it from observed experience. Intrinsic motivation \shortcite{singh04,linke20}, e.g., curiosity \shortcite{pathak17} can be used to deal with sparse rewards and encourage the agent to explore more. Imagination-based RL \shortcite{ha18,hafner20} is a solution that allows to train an agent, reducing at the same time the need for interaction with the environment. Hierarchical RL \shortcite{pateria21} allows to manage more complex problems by decomposing them into sub-tasks and working at different levels of abstraction. RL is not only used for training a single agent, but also in multi-agent scenarios~\shortcite{zhang21}.

\section{RL for Generative AI} \label{GRL}

In the following, we will discuss the state of the art in RL for generative learning considering three classes of solutions, which are summarized in Table \ref{tab:rl_approaches}: RL as an alternative solution for output generation with the goal of approximating outputs from a given domain of interest with high fidelity; RL as a way for generating output while maximizing an objective function which captures (additional) quantifiable properties or indicators at the same time; and, finally, RL as a way of embedding additional desired characteristics (such as value alignment) which cannot easily be captured by means of an objective function into the generative process.

\begin{table}[ht]
    \small
    \centering
    \begin{tabular}{cccc}
        \toprule
        Goal & Reward & Advantages & Limitations \\
        \midrule
        \begin{varwidth}{0.15\textwidth}
            Mere\\generation
        \end{varwidth} & 
        \begin{varwidth}{0.25\textwidth}
            \begin{itemize}
                \setlength\itemsep{0em}
                \item GAN's discriminative signal
                \item Log-likelihood of real or predicted targets
                \item Constraint satisfaction
        \end{itemize}
        \end{varwidth} & 
        \begin{varwidth}{0.25\textwidth}
            \begin{itemize}
                \setlength\itemsep{0em}
                \item Models domains defined by non-differentiable objectives
                \item Adapts GAN to sequential tasks
                \item Can implement RL strategies, e.g., hierarchical RL
            \end{itemize}
        \end{varwidth} &
        \begin{varwidth}{0.25\textwidth}
            \begin{itemize}
                \setlength\itemsep{0em}
                \item Learning without supervision is hard
                \item Pre-training can prevent an appropriate exploration
            \end{itemize}
        \end{varwidth} \\
        \cmidrule(r){1-4}
        \begin{varwidth}{0.15\textwidth}
            Objective\\maximization
        \end{varwidth} & 
        \begin{varwidth}{0.25\textwidth}
            \begin{itemize}
                \setlength\itemsep{0em}
                \item Test-time metrics
                \item Countable desired or undesired characteristics
                \item Distance-based measures
                \item Quantifiable properties
                \item Output of ML algorithms
        \end{itemize}
        \end{varwidth} & 
        \begin{varwidth}{0.25\textwidth}
            \begin{itemize}
                \setlength\itemsep{0em}
                \item Satisfies quantfiable requirements
                \item Optimizes a generator from a specific domain towards desirable sub-domains
                \item Reduces the gap between training and evaluation
            \end{itemize}
        \end{varwidth} &
        \begin{varwidth}{0.25\textwidth}
            \begin{itemize}
                \setlength\itemsep{0em}
                \item Not every desirable property is quantifiable or easy to get
                \item Goodhart's law
            \end{itemize}
        \end{varwidth} \\
        \cmidrule(r){1-4}
        \begin{varwidth}{0.15\textwidth}
            Improving\\not easily\\quantifiable\\characteristics
        \end{varwidth} & 
        \begin{varwidth}{0.25\textwidth}
            \begin{itemize}
                \setlength\itemsep{0em}
                \item Output of a model trained to reproduce human or AI feedback about non-quantifiable properties (e.g., helpfulness, appropriateness, creativity)
            \end{itemize}
        \end{varwidth} &
        \begin{varwidth}{0.25\textwidth}
            \begin{itemize}
                \setlength\itemsep{0em}
                \item Satisfies non-quantifiable requirements (for example, the alignment problem)
                \item Requires preferences between candidates instead of defining a mathematical measure of desired property
            \end{itemize}
        \end{varwidth} &
        \begin{varwidth}{0.25\textwidth}
            \begin{itemize}
                \setlength\itemsep{0em}
                \item Get user preferences is expensive
                \item Users might misbehave, disagree, or be biased
                \item Reward modeling is difficult
                \item Prone to jailbreaks out of alignment
            \end{itemize}
        \end{varwidth} \\
        \bottomrule
    \end{tabular}
    \caption{Summary of the three purposes for using RL with generative AI, considering the used rewards, their advantages, and their limitations.}
    \label{tab:rl_approaches}
\end{table}

\subsection{RL for Mere Generation}

\subsubsection{Overview}
\begin{figure}[ht]
  \centering
  \includegraphics[width=0.5\linewidth]{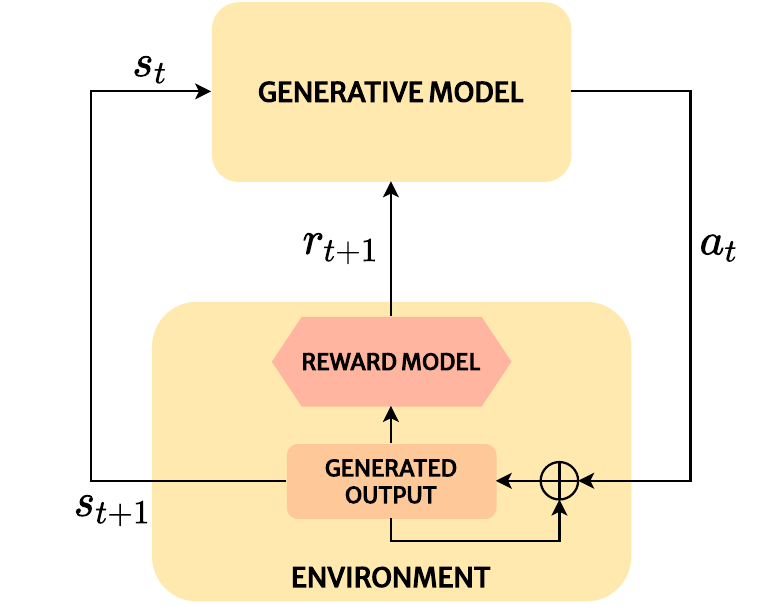}
  \caption[RL Framework for generative deep learning]{The reinforcement learning framework for generative modeling: at each timestep $t$, the Generative Model (i.e., the Agent) generates an action $a_t$ based on the current description of the generated output (i.e., current state) $s_t$, which updates the current description of the generated output to $s_{t+1}$, and receives a reward $r_{t+1}$ related to it.}
  \label{fig:rl_gdl}
\end{figure}

The simplest approach is RL for \textit{mere} generation, i.e., to train a generative model with the goal of approximating outputs from a given domain of interest as best as possible. Essentially, the objective function is used to replicate the behavior of the self-supervised learning loss used in traditional generative learning approaches, as the adversarial ones. In fact, due to its adherence to the formal framework of Markov decision processes \shortcite{sutton18}, RL can be used as a solution to the generative modeling problem in the case of sequential tasks \shortcite{bachman15}, e.g., text generation or stroke painting. 
The generative model plays the role of the agent. The current version of the generated output represents the state. For example, actions model how the state can be modified, e.g., which token\footnote{We use the term ``token'' to refer to any discrete element an unstructured data point can be broken into, independently the data source is in the form of text \shortcite<e.g.,>{radford19}, music \shortcite<e.g.,>{huang16}, image \shortcite<e.g.,>{ramesh21} and so on.} to be appended or which change applied to a picture. Finally, the reward is an indicator of the ``quality'' in terms of the generation of the output.
Figure \ref{fig:rl_gdl} summarizes the entire process.

It is possible to identify three fundamental design aspects: the implementation of the agent itself, e.g., diffusion model or transformer; the definition of the dynamics of the system, i.e., the transition between a state to another; the choice of the reward structure.
The first two depend on the task to be solved, e.g., music generation with LSTM composing one note after the other or painting with CNN superimposing subsequent strokes. The third one is instead responsible of the actual learning. While the reward can be structured so as to represent the classic supervised target, it also provides the designers with the opportunity of using a more diverse and complex set of reward functions, especially non-differentiable ones (which cannot be used in supervised learning due to the impossibility of computing their gradient for backpropagation).

The first example we consider is SeqGAN \shortcite{yu16}. Typically, GANs cannot be used for sequential tasks because the discriminative signal, i.e., whether the input looks real or not, is only available after the sequence is completed. SeqGAN circumvents this problem by using RL, which allows to learn from rewards received further in the future as well. Indeed, SeqGAN exploits the discriminative signal as the actual reward. The approach itself is based on a very simple policy approximation algorithm, namely REINFORCE \shortcite{williams92}. A similar approach is also used in MaskGAN \shortcite{fedus18}, where the generator learns with in-filling (i.e., by masking out a certain amount of words and then using the generator to predict them) through actor-critic learning \shortcite{sutton84}. Notably, hierarchical RL can also be used: for example, LeakGAN \shortcite{guo18} relies on a generator composed of a manager, which receives \textit{leaked} information from the discriminator, and a worker, which relies on a goal vector as a conditional input from the manager. Since SeqGAN might produce very sparse rewards, alternative strategies have been proposed. \shortciteA{shi18} suggest to replace the discriminator with a reward model learned with IRL on state-action pairs, so that the reward is available at each timestep (together with an entropy regularization term). A more complex state composed of a context embedding can also be used \shortcite{li19}. Instead, \shortciteA{li17} is based on a variation of SeqGAN: it uses Monte Carlo methods to get rewards at each timestep. In addition, the authors also suggest to alternate RL with a ``teacher'', i.e., the classic supervised training. This helps deal with tasks like text generation where the action space (i.e., the set of possible words or sub-words) is too large to be consistently explored using RL alone. Another solution to this problem is NLPO \shortcite{ramamurthy23}, which is a parameterized-masked extension of PPO \shortcite{schulman17} that restricts the action space via top-$p$ sampling, i.e., by only considering the smallest possible set of actions whose probabilities have a sum greater than $p$ \shortcite{holtzmann20}. TrufLL \shortcite{martin22} uses top-$p$ sampling as well; however, it restricts the action space by means of a pre-trained task-agnostic model \textit{before} applying policy gradient with PPO. Similarly, ColdGAN \shortcite{scialom20} forces the sampling of a SeqGAN-like generator to be close to the distribution modes by selecting actions with top-$p$ sampling and low temperature \shortcite{holtzmann20} and training the generator via importance sampling \shortcite{precup00}. Finally, \shortciteA{lamprier22} propose to substitute a top-$p$ sampling strategy with a cooperative one based on the use of Monte Carlo Tree Search structure, which is evaluated by the discriminator; again, the generator is trained via importance sampling.

Another reason to use RL is to take advantage of its inherent properties. For example, GOLD \shortcite{pang21} is an algorithm that substitutes self-supervised learning with off-policy RL and importance sampling. It uses real demonstrations, which are stored in a replay buffer; the reward corresponds to either the sum or the product of the action probabilities over the sampled trajectories, i.e., of each single real token according to the model. While it can be considered close to a self-supervised approach, off-policy RL with importance sampling allows up-weighting actions with high (cumulative) return and actions preferred by the current policy, encouraging to focus on in-distribution examples.

RL is also an effective solution for learning in domains in which a differentiable objective is difficult or impossible to define. RL-Duet \shortcite{jiang20} is an algorithm for online accompaniment generation. Learning how to produce musical notes according to a given context is a complex task: RL-Duet first learns a reward model that considers both inter-part (i.e., with counterpart) and intra-part (i.e., on its own) harmonization. Such model is composed by an ensemble of networks trained to predict different portions of music sheets (with or without human counter-part, and with or without machine context). Then, the generative system is trained to maximize this reward by means of an actor-critic architecture with generalized advantage estimator \shortcite<GAE;>{schulman16}. CodeRL \shortcite{le22} performs code generation through a pre-trained model and RL. In particular, the model is fine-tuned with policy gradient in order to maximize the probability of passing unit tests: it receives a (sparse) reward quantifying if (and how) the generated code has passed the test for the assigned task. In addition, a critic learns a (dense) signal to predict the compiler output. The model is then trained to maximize both signals considering a baseline obtained with a greedy decoding strategy. In order to obtain a denser and more informative reward, PPOCoder \shortcite{shojaee23} also considers three additional signals: a syntactic matching score based on the Abstract Syntax Tree of the generated code; a semantic matching score based on the data-flow graph; and a Kullback-Leibler (KL) penalty to prevent the model from deviating considerably from its pre-trained version. The sum of these four signals is then optimized via PPO.

Another interesting application area is painting. \shortciteA{xie12} suggest to model stroke painting as a Markov Decision Process, where the state is the canvas, and the actions are the brushstrokes performed by the agent. Rewards calculated considering the location and inclination of the strokes are then used to train the agent. For instance, Doodle-SDQ \shortcite{zhou18} fine-tunes a pre-trained sketcher with Double DQN \shortcite{vanhasselt16} and a reward that is calculated by evaluating how well a sketch reproduces a target image at pixel, movement, and color levels. \shortciteA{huang19} use a discriminator trained to recognize real canvas-target image pairs to derive a corresponding reward. Instead, \shortciteA{singh21} train a painting policy that operates at two different levels: foreground and background. Each of them uses a discriminator; in addition, they adopt a focus reward measuring the degree of indistinguishability of two object features. On the other hand, Intelli-Paint \shortcite{singh22} is based on four different types of rewards, which are used to learn a painting policy with deep deterministic policy gradient \shortcite<\fake{DDPG};>{lillicrap16} based on a discriminator signal on canvas-image pairs, two penalties for the color and position of consecutive strokes, and the same semantic guidance proposed by \shortciteA{singh21}. Finally, RL has also been used for collage artwork. \shortciteA{lee23c} propose an RL-based method with the goal of composing different elements (such as newspaper or texture cuts) in order to obtain an output that resembles a target picture. The state is composed by the canvas, the target image, and randomly (or value-based) sampled material; the action determines which region of the material to cut and where to paste it on current canvas; and the reward is the amount of similarity change between consecutive timesteps (where the similarity between the canvas and the target image is computed by a WGAN-GP discriminator \shortcite{gulrajani17} trained in parallel to discriminate between target-target and target-canvas pairs). A model-based soft actor-critic \shortcite<SAC;>{haarnoja19} is then used to optimize the reward minus a penalty for each timestep in order to teach the agent to complete the tasks with the minimum number of actions.

\subsubsection{Discussion}
RL can represent an alternative method for deriving generative models, especially if the target loss is non-differentiable. 
It allows for the adaptation of known generative strategies, e.g., GANs, to tasks for which traditional techniques are not suitable, e.g., in text generation. In addition, it can be applied to domains in which feasibility and correctness (e.g., running code as above) are essential dimensions to consider. In other words, RL can train a generative model to produce observations that appear to have been drawn from the domain of interest even when such domain cannot be modeled by means of generative functions and corresponding differentiable losses.
RL can also be used to derive more complex generative strategies (e.g., through hierarchical RL) and to reduce the model dependence on training data, which might have an impact on copyright issues \shortcite{franceschelli22,henderson23}.

It is possible to identify some limitations of the proposed solution.
Learning without supervision is a hard task, especially when the reward is sparse. This is very likely to happen for sequence generation, such as (long) text or music, where the reward is available only at the last timestep. In addition to the aforementioned techniques for obtaining a denser reward, a potential solution might consist in considering an intrinsic reward \shortcite{aubret19} as an additional learning signal, in order to encourage exploration as well.
Moreover, the action space can be very large (potentially orders of magnitude larger than those of standard RL problems, \shortciteR{ammanabrolu20}), especially for text generation. Ensuring a sufficient exploration of all possible actions while still exploiting the most promising ones to collect higher rewards is one of the key problems in RL. Starting with some prior knowledge about the possible best actions for different situations might be necessary for fast convergence.
For this reason, pre-trained generative models are selected for this task. This can cause the agent to initially focus on highly probable tokens, increasing their associated probabilities and, because of that, failing to explore different solutions (i.e., by only moving the probability mass of the already most probable tokens) \shortcite{choshen20}. These problems can be avoided through variance reduction techniques (e.g., incorporating baselines and critics) and exploration strategies \shortcite{kiegeland21}.

\subsection{RL for Objective Maximization}
\subsubsection{Overview}
RL can be formalized and studied as an objective maximization problem. In this subsection we will discuss how this type of formalization can be applied to generative AI.
Since RL allows us to use any non-differentiable function for modeling the rewards, it could be the case that simply replicating the behavior of a self-supervised learning loss is not the optimal solution. For example, \shortciteA{ranzato16} point out the mismatch between how deep learning models are trained (i.e., on differentiable losses) and how they are commonly evaluated (i.e., on non-differentiable metrics): an emerging line of research is focusing on the use of non-differentiable metrics as reward functions for generative learning capturing a variety of requirements and constraints.


RL for quantity maximization has been mainly adopted in text generation, especially for dialogue and translation. In addition to exposure bias mitigation, it allows for replacing classic likelihood-based losses with metrics used at inference time. A pioneering work is the one by \shortciteA{ranzato16}, where RL is adopted to directly maximize BLEU \shortcite{papineni02} and ROUGE \shortcite{lin04} scores. To deal with the size of the action space, the authors introduce MIXER, a variant of REINFORCE algorithm that uses incremental learning (i.e., an algorithm based on an optimal pre-trained model according to ground truth sequences) and combines reward maximization with classic cross-entropy loss by means of an annealing schedule. In this way, the model starts with preexisting knowledge, which is preserved through the classic loss, while aiming at exploring alternative but still probable solutions, which should increase score at test time. A similar approach is also used by Google's neural machine translation system \shortcite{wu16}. BLEU score is used as the reward, while fine-tuning a pre-trained neural translator with a mixed maximum likelihood and expected reward objective. \shortciteA{bahdanau17} consider an actor-critic algorithm for machine translation, with the critic conditioned on the target text, and the pre-trained actor fine-tuned with BLEU as the reward. \shortciteA{paulus18} suggest to learn to perform text summarization by using self-critical policy training \shortcite{rennie17}, where the reward associated with the action that would have been chosen at inference time is used as baseline. ROUGE score is considered as the reward, and linearly mixed with teacher forcing \shortcite{williams89}, i.e., classic supervised learning. Scores alternative to ROUGE have been proposed as well, e.g.,  ROUGESal and Entail both described by \shortciteA{pasunuru18}. The former up-weights the salient sentences or words detected via a key-phrase classifier. The latter rewards logically-entailed summaries through an entailment classifier. They are then used alternatively in subsequent mini-batches to train a Seq2Seq model \shortcite{sutskever14} by means of REINFORCE. Finally, \shortciteA{zhou17} consider BLEU score to train a dialogue system on top of collected human interactions with offline RL. An additional dialogue-level reward function (measuring the number of proposed API calls) is also used. Recently, the RL4LM library \shortcite{ramamurthy23} started offering many of these metrics as rewards, thus facilitating their use for LM training or fine-tuning. Different families of solutions are considered, i.e., $n$-grams overlapping such as ROUGE, BLEU, SacreBLEU \shortcite{post18} or METEOR \shortcite{lavie07}; model-based methods such as BertScore \shortcite{zhang20} or BLEURT \shortcite{sellam20}; task-specific metrics; and perplexity. Notably, RL4LM also allows to balance such metrics with a KL-divergence minimization with respect to a pre-trained model.

Test-time metrics are not the only quantities that can be maximized through RL. For example, \shortciteA{lagutin21} suggest considering the count of 4-gram repetitions in the generated text, to reduce the likelihood of undesirable results.
The combination of these techniques and classic self-supervised learning helps learn both \textit{how to write} and \textit{how not to write}. \shortciteA{li16} train a Seq2Seq model for dialogue by rewarding conversations that are informative (i.e., which avoid repetitions), interactive (i.e., which reduce the probability of answers like ``I don't have any idea'' that do not encourage further interactions), and coherent (i.e., which are characterized by high mutual information with respect to previous parts of the conversation). Sentence-level cohesion (i.e, compatibility of each pair of consecutive sentences) and paragraph-level coherence (i.e., compatibility among all sentences in a paragraph) can be achieved by maximizing the cosine similarity between the encoded version of the relative text, with the encoders trained so that the entire discriminative models are able to distinguish between real and generated pairs \shortcite{cho19}. A distance-based reward can instead guide a plot generator towards reaching desired goals. \shortciteA{tambwekar19} train an agent working at event level (i.e., a tuple with the encoding of a verb, a subject, an object, and a fourth possible noun) with REINFORCE to minimize the distance between the generated verb and the goal verb. Other domain-specific rewards are used by \shortciteA{yi18}, where two distinct generative models produce poetry by maximizing fluency (i.e., MLE on a fixed language model), coherence (i.e., mutual information), meaningfulness (i.e., TF-IDF), and overall quality from a learned classifier. In addition, the two models also learn from each other: the worst performing can be trained on the output produced by the other one, or its distribution can be modified in order to better approximate the other.

Another popular technique is hierarchical RL: it allows optimization of quantifiable objectives even when they work at a different level of abstraction with respect to the generative model. For example, \shortciteA{peng17} uses it to design a dialogue system able to perform composite tasks, i.e., sets of subtasks that need to be performed collectively. A high-level policy, trained to maximize an extrinsic reward directly provided by the user after each interaction, selects the sub-tasks. Then, ``primitive'' actions to complete the given sub-task are chosen according to a lower-level policy. A global state tracker on cross-subtask constraints is employed in order to provide the RL model with an intrinsic reward measuring how likely a particular subtask will be completed. Finally, ILQL \shortcite{snell23} learns a state-action and a state-value function that are used to perturb a fixed LLM, rather than directly fine-tuning the model itself. This allows to preserve the capabilities of the given pre-trained language model, while still maximizing a specific utility function.

While text generation is one of the areas that have attracted most of the attention of researchers and practitioners in the past years, RL with quantity maximization has been applied to other sequential tasks as well. An important line of research \shortcite{jaques16,jaques17a,jaques17b} consists of fine-tuning a pre-trained sequence predictor with imposed reward functions, while preserving the learned properties from data. For instance, a pre-trained note-based RNN can represent the starting point for the Q-network in DQN. A reward given by the probability of the chosen token according to the original model (or based on the inverse of the KL divergence) and one based on music theory rules (e.g., that all notes must belong to the same key) are used to fine-tune the model. Another possibility is to extend SeqGAN to domain-specific reward maximization, as in ORGAN \shortcite{guimaraes17}. ORGAN linearly combines the discriminative signal with desired objectives, also dividing the reward by the number of repetitions made, in order to increase diversity in the result. Music generation can then be performed by considering tonality and ratio of steps as rewards; solubility, synthesizability and drug-likenesses are instead adopted to perform molecule generation as a sequential task, i.e., by considering a string-based representation of molecules (by means of SMILES language, \shortciteR{weininger88}). While the original work considered RNN-based models, transformer architectures can be used as well \shortcite{li22}. 

Molecular generation is indeed one of the most explored task at the intersection between RL and generative AI. While MolGAN \shortcite{decao18} adapts ORGAN to graph-based generative models \shortcite{li18} to directly produce molecular structures, the majority of research focuses on simplified molecular-input line-entry system (SMILES) textual notation \shortcite{weininger88}, so as to leverage the recent advancements in text generation. ReLeaSe \shortcite{popova18} fine-tunes a pre-trained generator to maximize physical, biological, or chemical properties (learned by a reward model). \shortciteA{olivecrona17} propose to fine-tune a pre-trained generator with REINFORCE so as to maximize a linear combination of a prior likelihood (to avoid catastrophic forgetting) and a user-defined scoring function (e.g., to match a provided query structure or to have predicted biological activity). REINVENT \shortcite{blaschke20} also avoids to generate molecules the model already produced through a memory that keeps track of the good scaffoldings generated so far. \shortciteA{atance22} adopt REINVENT for the graph-based deep generative model GRAPHINVENT \shortcite{mercado21} in order to directly obtain molecules that maximize desired properties, e.g., pharmacological activity. Instead, GENTRL \shortcite{zhavoronkov19} generates kinase inhibitors relying on a variational autoencoder to reduce molecules to continuous latent vectors. Then REINFORCE is used to teach the decoder how to maximize three properties learned through self-organizing maps: activity of compounds against kinases; closeness to neurons associated with DDR1 inhibitors within the whole kinase map; and novelty of chemical structures. The average reward for the produced batch is assumed as a baseline to reduce variance. Notably, RL is used here for single-step generation (i.e., by means of a contextual bandit). \shortciteA{gaudin19} propose to generate molecules maximizing their partition coefficient without any pre-training by working with a simplified language \shortcite{krenn20}; \shortciteA{thiede22} suggest to use intrinsic rewards to better explore its solution space. GCPN \shortcite{you18} trains a graph-CNN to optimize domain-specific rewards and an adversarial loss (from a GAN-like discriminator) through PPO. Other tasks have been investigated as well. \shortciteA{nguyen22} merge GAN and actor-critic in order to obtain a generator capable of producing 3D material microstructures with desired properties. \shortciteA{han20} use DDPG to train an agent to design buildings (in terms of shape and position) so as to maximize several signals related to the performance and aesthetics of the generated block, e.g., solar exposure, collision, and number of buildings.

Finally, the use of techniques based on objective maximization can also be effective for image generation. Denoising Diffusion Policy Optimization \shortcite<\fake{DDPO};>{black23} can train or fine-tune a denoising diffusion model to maximize a given reward. It considers the iterative denoising procedure as a Markov Decision Process of fixed length. The state contains the conditional context, the timestep, and the current image; each action represents a denoising step; and the reward is only available for the termination state, when the final, denoised image is obtained. DDPO has therefore been used to learn how to generate images that are more compressed or uncompressed, by minimizing or maximizing JPEG compression; more aesthetically pleasing, by maximizing LAION score\footnote{\url{https://laion.ai/blog/laion-aesthetics/}}; or more prompt-aligned, by maximizing the similarity between the embeddings of prompt and generated image description. Improving the aesthetics of the image while preserving the text-image alignment has also been done at the prompt level \shortcite{hao22}. A language model that given human input provides an optimized prompt can be trained with PPO to maximize both an aesthetic score (from an aesthetic predictor) and a relevance score (as CLIP embedding similarity) of the image generated from the given prompt.  

\subsubsection{Discussion}

Reinforcement learning for objective maximization opens up several new possibilities: generators can be adapted for particular domains or for specific problems; they can be built for tasks difficult to model through differentiable functions; and pre-trained models can be fine-tuned according to given requirements and specifications.
Essentially, RL is not used only for mere generation, since it also allows more specific, goal-oriented generative modeling
: instead of training a generator to produce \textit{correct, reasonable examples} for the domain of interest, the goal is to derive \textit{the best possible examples} according to some specific target functions.
Any desired and quantifiable property can now be set as reward function, thus in a sense ``teaching'' a model how to achieve it.  While research has focused its attention on sequential tasks like text or music generation, other domains might be considered as well. As shown by \shortciteA{zhavoronkov19}, tasks not requiring multiple generative steps can be performed simply by reducing the RL problem to a contextual bandit one. In this way, RL can be considered as a technique for specific sub-domains, in a manner similar to neural style transfer \shortcite{gatys16} or prompt engineering \shortcite{liu22b}.

We can identify possible drawbacks as well. Reinforcement learning has typically a very high computational cost \shortcite{ceron21}, due to the number of iterations required to converge. In addition, certain desired properties (e.g. harmlessness or appropriateness) can be difficult to quantify, or the related measures can be expensive to compute, especially at run-time. This can lead to excessive computational time for training. While offline RL might alleviate this problem, it would require a collection of evaluated examples, thus eliminating the advantage of not needing a dataset and increasing the risk of exposure bias. Finally, a fundamental issue arises from using test-time metrics as objective functions: how should we evaluate the model we derive? In fact, according to the empirical Goodhart's Law \shortcite{goodhart75}, ``when a measure becomes a target, it ceases to be a good measure''. New metrics are then required, and a gap between training objective and test score might be inevitable.

\subsection{RL for Improving Not Easily Quantifiable Characteristics}
\subsubsection{Overview}
While test-time metrics as objectives reduce the gap between training and evaluation, they not always correlate with human judgment \shortcite{chaganty18}. In these cases, using such metrics would not help obtain the desired generative model. Moreover, there might be certain qualities that do not have a correspondent metric because they are subjective, difficult to define, or, simply, not quantifiable. Typically users only have an implicit understanding of the task objective, and, therefore, a suitable reward function is almost impossible to design: this problem is commonly referred to as the \textit{agent alignment problem} \shortcite{leike18}. 

One of the most promising directions is reward modeling, i.e., learning the reward function from interaction with the user and then optimizing the agent through RL over such function. In particular, Reinforcement Learning from Human Feedback \shortcite<RLHF;>{christiano17} allows to use human feedback to guide policy gradient methods. A reward model is trained to associate a reward to a trajectory thanks to human preferences (so that the reward associated with the preferred trajectory is higher than that associated with the others). In parallel, a policy is trained by means of this signal using a policy gradient method, while the trajectories collected at inference time are used to obtain new human feedback to improve the model. \shortciteA{ziegler19} apply RLHF to text continuation, e.g., to write positive continuations of text summaries. A pre-trained language model is used to sample text continuations, which are then evaluated by humans; a reward model is trained over such preferences; and finally, the policy is fine-tuned using KL-PPO \shortcite{schulman17} in order to maximize the signal provided by the reward model. A KL penalty is used to prevent the policy moving too far from its original version. Notably, these three steps can be performed once (offline case) or multiple times (online case). 

Similarly, \shortciteA{stiennon20} use RLHF to perform text summarization. The following three steps are repeated one or more times: human feedback collection, during which for each sampled reddit post different summaries are generated, and then human evaluators are asked to rank them; reward model training on such preferences; policy training with PPO with the goal of maximizing the signal from the reward model (still using a KL penalty). \shortciteA{wu21} propose to summarize entire books with RLHF by means of recursive task decomposition, i.e., by first learning to summarize small sections of a book, then summarizing those summaries into higher-level summaries, and so on. In this way, the size of the texts to be summarized is smaller. This is more efficient in terms of generative modeling and human evaluation, since the samples to be judged are shorter.
InstructGPT \shortcite{ouyang22} fine-tunes GPT-3 \shortcite{brown20} with RLHF so that it can follow written instructions. With respect to \shortciteA{stiennon20}, demonstrations of desired behavior are first collected from humans and used to fine-tune GPT-3 before actually performing RLHF. Then, a prompt is sampled and multiple model outputs are generated, with a human labeler ranking them. Such rankings are finally used to train the reward model. The latter is then utilized (together with a KL penalty) to train the actual RL model with PPO. In particular, this procedure is adopted in ChatGPT and GPT-4 \shortcite{openai23}, which are fine-tuned in order to be aligned with human judgment.

Although all these methods consider human feedback regarding the ``best'' output for a given input (with ``best'' generally meaning appropriate, factual, respectful, or qualitative), more specific or different criteria are also used. \shortciteA{bai22a} consider human preferences for helpfulness and harmlessness. Sparrow \shortcite{glaese22} is trained to be helpful, correct, and harmless, with the three criteria judged separately so that three more efficient rule-conditional reward models are learned. In addition, the model is trained to query the web for evidence supporting provided facts; and again RLHF is used to obtain human feedback about the appropriateness of the collected evidence. Finally, \shortciteA{pardinas23} use RLHF to fine-tune GPT-2 to learn how to write \textit{haikus} maximizing the relevance to the provided topic, self-consistency, creativity, form, and avoiding toxic content through human feedback. In addition to text, RLHF has been used to better align text-to-image generation with human preferences. After collecting user feedback about text-image alignment, a reward model is learned to approximate such feedback, and its output is used to weight the classic loss function of denoising diffusion models \shortcite{lee23}. On the contrary, DPOK \shortcite{fan23} directly applies online reinforcement learning for fine-tuning text-to-image diffusion models, which are optimized using a learned reward model from human feedback \shortcite{xu23} and a KL regularization with respect to the pre-trained model.

While very effective, RLHF is not the only existing approach. When human ratings are available in advance for each piece of text, a reward model can be trained offline and then used to fine-tune an LLM \shortcite{bohm19}. Such a reward model can also be combined with classic MLE to effectively train a language model \shortcite{kreutzer18} or used to pre-pend reward tokens to generated text, forming a replay buffer suitable for online, off-policy algorithms to unlearn undesirable properties \shortcite{lu22}. Alternatively, A-LoL \shortcite{baheti23} adopts offline policy gradient with a single-action step assumption (i.e., the entire sequence is a single action) to optimize for pre-trained, sequence-level reward models; in order to improve learning efficiency, it filters out data points with negative advantages, with the critic based on a frozen reference LLM. Since human ratings might be inaccurate, \shortciteA{nguyen17} suggest to simulate them by applying perturbations on automatically generated scores. Alternatively, the provided dataset of scored text allows for batch (i.e., offline) policy gradient methods to train a chatbot \shortcite{kandasamy17}. A very similar approach is also followed by \shortciteA{jaques20}, where offline RL is used to train a dialogue system on collected conversations (with relative ratings) filtered to avoid learning misbehavior. Other strategies can be implemented as well. RELIS \shortcite{gao19} relies on a learned reward model from human-provided judgment as the other systems discussed above; however, such reward model is used to optimize a policy directly at inference time for the provided text. Instead of training a policy over multiple inputs and then exploiting it at inference time, it trains a different policy for each required input.

Another possibility is to use AI feedback instead of, or in addition to, the human one. Constitutional AI \shortcite{bai22b} is a method to train a non-evasive and ``harmless'' AI assistant without any human feedback, only relying on a \textit{constitution} of principles to follow\footnote{While the selection of precepts to be included in the original ``constitution'' is defined by the researchers at Anthropic, a follow-up project called Collective Constitutional AI \shortcite{anthropic23} involves the participation of humans  for crowd-sourcing the underlying principles by means of Polis \shortcite{small21}, which is a platform for running online deliberative processes augmented by machine learning algorithms.}. In a first supervised stage, a pre-trained LLM is used to generate responses to prompts, and then to iteratively correct them to satisfy a set of principles; once the response is deemed acceptable, it is used to fine-tune the model. Then, RLHF is performed, with the only difference that feedback is provided by the model itself and not by humans. RLAIF \shortcite{lee23b} completely replaces human preferences with preferences from an off-the-shelf LLM for text summarization. The desired overall behavior is induced by careful prompting. \shortciteA{liu22a} use RL to fine-tune a Seq2Seq model to generate knowledge useful for a generic QA model. This is first re-trained on knowledge generated with GPT-3 (which is prompted asking to provide the knowledge required to answer a certain question). Then, RL is used to fine-tune the model so as to maximize an accuracy score using knowledge generated by the model itself as a prompt. To avoid catastrophic forgetting, a KL penalty (with respect to the initial model) is introduced. RNES \shortcite{wu18} is instead a method to train an extractive summarizer (i.e., a component that selects which sentences of a given text should be included in its summary) using a reward based on coherence. A model is trained to identify the appropriate next sentence composing a coherent sentence pair; then, such a signal is used to obtain immediate rewards while training the agent (with ROUGE as the reward for the final composition). Finally, \shortciteA{su16} propose to limit requests for human feedback to cases in which the learned reward model is uncertain.

\subsubsection{Discussion}
Reward modeling, i.e., learning the reward function from interaction with users, introduces a great level of flexibility in RL for generative AI. Generative models can be trained to produce content that humans consider appropriate and of sufficient quality, by aligning them with their preferences.
This is useful and in many situations essential: in fact a quantifiable measure might not exist or information to derive it might be hard to obtain.
This methodology has already shown its intrinsic value in obtaining accurate, helpful, and useful text. In the same way, these techniques can be applied to other domains in which desired qualities are difficult to quantify or hard to express in a mathematical form, e.g., aesthetically pleasant or personalized (multimodal) content or creative artifacts \shortcite{franceschelli23}. A recap on covered applications is reported in Table \ref{tab:applications}.

RLHF has proven to be a highly effective approach. However, it suffers from several open problems \shortcite{casper23}. For example, getting user feedback can be incredibly expensive. Moreover, the users might misbehave, whether on purpose or not, be biased, or disagree within each other \shortcite{fernandes23}. Also, they might not correctly represent the population of end users or marginalized categories; and comparison-based feedback may not correlate with the desirability of responses \shortcite{casper23}. For these reasons, other techniques for modeling preferences might be considered. If human ratings are available in advance, a reward model can be derived from them and used in offline mode. Using AI itself to provide feedback is also an option; notably, AI-based feedback is also used outside the RL paradigm, e.g., to provide verbal feedback to be appended to prompts \shortcite{shinn23} or to collaborate with other LLMs at inference time \shortcite{dong23,du23}. In addition, other techniques such as IRL or cooperative IRL \shortcite{hadfieldmenell16} can be applied to induce a reward model from human demonstrations.

Reward modeling can be problematic as well. Reducing the diversity of society to a single reward function might cause the majority views to disproportionately prevail \shortcite{feffer23}. In addition, seemingly well-performing preference-based reward models might fail to generalize to out-of-distribution states \shortcite{tien23}, thus being prone to reward hacking (i.e., optimizing an imperfect proxy reward function that leads to poor performance according to the true reward function, \shortciteR{skalse22}).  For these reasons, recent work has focused on eliminating the need for a reward model at all (e.g., \shortciteR{rafailov23,song23}).

Finally, \shortciteA{wolf23} show that, even if aligned, LLMs can still be prompted in ways that lead to undesired behavior. In particular, ``jailbreaks'' out of alignment can be obtained via single prompts, especially when asking the model to simulate malicious personas \shortcite{deshpande23}. This is more likely to happen in the case of aligned models rather than of non-aligned ones because of the so-called \textit{waluigi effect}: by learning to behave in a certain way, the model also learns its exact opposite \shortcite{nardo23}. More advanced approaches would be required to mitigate this problem and completely prevent certain undesired behaviors.

\begin{table}[ht]
    \tiny
    \centering
    \begin{tabular}{ |p{0.17\textwidth}|p{0.44\textwidth}|p{0.07\textwidth}|p{0.20\textwidth}|  }
        \hline
        Application& Reward &RL Type &Example \\
        \hline
        \rowcolor{yellow!50} \multicolumn{4}{|l|}{Chemistry} \\
        \rowcolor{oran!0} & Discriminator + chemical properties &P & \shortcite{decao18} \\ \rowcolor{oran!0} & Pharmacological activity + prior likelihood &P & \shortcite{atance22} \\ \rowcolor{oran!0} & Adversarial loss + desired properties &P & \shortcite{you18} \\ \rowcolor{oran!0} \multirow{-4}{*}{Molecule (graph)} & Novelty + utility of inhibitors &CB & \shortcite{zhavoronkov19} \\
        \rowcolor{oran!50} & Discriminator + chemical properties &P & \shortcite{guimaraes17} \\ \rowcolor{oran!50} & Learned desired properties &P & \shortcite{popova18} \\ \rowcolor{oran!50} & Desired property + prior likelihood &P & \shortcite{olivecrona17} \\ \rowcolor{oran!50} & As above + penalty for repetitions &P & \shortcite{blaschke20} \\ \rowcolor{oran!50} & Partition coefficient &TD & \shortcite{gaudin19} \\ \rowcolor{oran!50} \multirow{-6}{*}{Molecule (text)} & Desired property + intrinsic reward &P & \shortcite{thiede22} \\
        \hline
        \rowcolor{yellow!50} \multicolumn{4}{|l|}{Computer Vision} \\
        \rowcolor{oran!0} Collage & Discriminator on canvas-target pairs + length penalty &P & \shortcite{lee23c} \\
        \rowcolor{oran!50} Image & Compression or aesthetic or prompt alignment &P & \shortcite{black23} \\
        \rowcolor{oran!0} & Pixel, movement, color reproduction &TD & \shortcite{zhou18} \\ \rowcolor{oran!0} & Discriminator on canvas-target pairs &P & \shortcite{huang19} \\ \rowcolor{oran!0} &Background vs foreground + focus &P & \shortcite{singh21} \\ \rowcolor{oran!0} \multirow{-4}{*}{Stroke painting} & Two above + adjacent color/position &P & \shortcite{singh22} \\
        \rowcolor{oran!50} & RLHF on text-image alignment &RWCE &\cite{lee23} \\ \rowcolor{oran!50} \multirow{-2}{*}{Text-to-image} & Learned reward model from human feedback &P &\cite{fan23} \\
        \hline
        \rowcolor{yellow!50} \multicolumn{4}{|l|}{Design} \\
        \rowcolor{oran!0} Building &Performance and aesthetic metrics &P &\shortcite{han20} \\
        \rowcolor{oran!50} Microstructure &Adversarial loss + target properties &P &\shortcite{nguyen22} \\
        \hline
        \rowcolor{yellow!50} \multicolumn{4}{|l|}{Music} \\
        \rowcolor{oran!0} Accompaniment &Log-likelihood for pre-trained models &P & \shortcite{jiang20} \\
        \rowcolor{oran!50} & Discriminator signal &P & \shortcite{yu16} \\ \rowcolor{oran!50} & Music theory rules + log-likelihood for original model &TD & \shortcite{jaques16} \\ \rowcolor{oran!50} \multirow{-3}{*}{Music} & Discriminator signal + tonality + ratio of steps &P & \shortcite{guimaraes17} \\
        \hline
        \rowcolor{yellow!50} \multicolumn{4}{|l|}{Natural language} \\
        \rowcolor{oran!0} & Discriminator signal at each $t$ through MC methods &P & \shortcite{li17} \\ \rowcolor{oran!0} & Discriminator signal at each $t$ through IRL &P & \shortcite{shi18} \\ \rowcolor{oran!0} & Repetitive or useless answer penalty + mutual information&P  & \shortcite{li16} \\ \rowcolor{oran!0} & Reward from user + likelihood of sub-task completion &HP & \shortcite{peng17} \\ \rowcolor{oran!0} & BLEU + number of proposed API calls &OffP & \shortcite{zhou17} \\ \rowcolor{oran!0} & RLHF &P & \shortcite{ouyang22} \\ \rowcolor{oran!0} & RLHF on helpfulness and harmlessness &P & \shortcite{bai22a} \\ \rowcolor{oran!0} & RLHF on helpfulness, harmlessness and correctness &P & \shortcite{glaese22} \\ \rowcolor{oran!0} & AI feedback based on a constitution of principles &P & \shortcite{bai22b} \\ \rowcolor{oran!0} & Collected human ratings &OffP & \shortcite{kandasamy17}\\ \rowcolor{oran!0} & Learned reward model of human ratings &TD & \shortcite{su16} \\ \rowcolor{oran!0} \multirow{-12}{*}{Chatbot} & Learned sequence-level reward model of human preferences &OffP & \shortcite{baheti23} \\ 
        \rowcolor{oran!50} &Reward model from human ratings &TD &\shortcite{gao19} \\ \rowcolor{oran!50} \multirow{-2}{*}{\shortstack[l]{Extractive \\summarization}} &Coherence ratings + ROUGE &P &\shortcite{wu18} \\
        \rowcolor{oran!0} & Discriminator signal &P & \shortcite{fedus18} \\ \rowcolor{oran!0} & Sum or product of log-likelihood of tokens from target text &OffP & \shortcite{pang21} \\ \rowcolor{oran!0} & 4gram repetition penalty + log-likelihood of target output &P & \shortcite{lagutin21} \\ \rowcolor{oran!0} & Discriminator signals on coherence and cohesion &P & \shortcite{cho19} \\ \rowcolor{oran!0} \multirow{-5}{*}{Generic text} & Specific utility function to maximize at inference time &TD & \shortcite{snell23} \\
        \rowcolor{oran!50} Knowledge & Accuracy score + kl penalty &P &\cite{liu22a} \\
        \rowcolor{oran!0} & BLEU + log-likelihood of target output &P & \shortcite{ranzato16}\\ \rowcolor{oran!0} & BLEU &P & \shortcite{bahdanau17} \\ \rowcolor{oran!0} & Implicit task-based feedback from users &P & \shortcite{kreutzer18} \\ \rowcolor{oran!0} \multirow{-4}{*}{Machine translation} & Perturbed predicted human ratings &CB & \shortcite{nguyen17} \\
        \rowcolor{oran!50} Plot & Generated vs target verbs distance &P & \shortcite{tambwekar19} \\
        \rowcolor{oran!0} Prompt optimization & Aesthetic score + CLIP similarity &P & \shortcite{hao22} \\
        \rowcolor{oran!50} & Discriminator signal &P & \shortcite{yu16} \\ \rowcolor{oran!50} & Fluency + coherence + meaningfulness + quality &P & \shortcite{yi18} \\ \rowcolor{oran!50} \multirow{-3}{*}{Poetry} & RLHF on relevance, consistency, creativity, form, toxicity &P & \shortcite{pardinas23} \\
        \rowcolor{oran!0} Text continuation & RLHF &P & \shortcite{ziegler19} \\
        \rowcolor{oran!50} & ROUGE + log-likelihood of target output &P & \shortcite{paulus18} \\ \rowcolor{oran!50} & ROUGESal + Entail &P & \shortcite{pasunuru18} \\ \rowcolor{oran!50} & RLHF &P & \shortcite{stiennon20}\\ \rowcolor{oran!50} & Reward model trained on human ratings &TD & \shortcite{bohm19} \\ \rowcolor{oran!50} \multirow{-5}{*}{Text summarization} & RLAIF &P & \shortcite{lee23b} \\
        \hline
        \rowcolor{yellow!50} \multicolumn{4}{|l|}{Programming} \\
        \rowcolor{oran!0} Code & Result of unit tests &P & \shortcite{le22} \\
        \hline
    \end{tabular}
    \caption{Summary of all the applications covered by past research in RL for generative AI, with the considered rewards and the relative references. Type of algorithms used: On-\textbf{P}olicy; \textbf{Off-P}olicy; \textbf{T}emporal-\textbf{D}ifference; \textbf{C}ontextual \textbf{B}andit; \textbf{H}ierarchical \textbf{P}olicy; \textbf{R}eward-\textbf{W}eighted \textbf{C}ross-\textbf{E}ntropy.}
    \label{tab:applications}
\end{table}

\section{Conclusion} \label{conclusion}

Reinforcement learning for generative AI has attracted huge attention after the recent breakthroughs in the area of foundation models and, in particular, large-scale language models. In this survey, we have investigated the state of the art, the opportunities and the open challenges in this fascinating area. First, we have discussed RL for classic generation, where RL simply provides a suitable framework for domains that cannot be modeled by means of a well-defined and differentiable objective, also reducing exposure bias. Then, we have considered RL for quantity maximization, where RL is used to teach a commonly pre-trained model how to maximize a numerical property. This closes the gap between what the model is optimized for and how it is evaluated, but also to search for particular characteristics and sub-domains. Finally, we have analyzed RL for non-easily quantifiable characteristics, where RL is used for aligning it with human requirements and preferences that are not easily expressed in a mathematical form.

Since non-differentiable functions can be used as target objectives, RL allows for a broader adoption of generative modeling, taking into consideration a wide range of objectives, requirements and constraints. Current and emerging solutions are characterized by the integration of a variety of RL mechanisms, such as IRL, hierarchical RL or intrinsic motivation, just to name a few.
On the other hand, the use of RL for generative AI introduces the problem of balancing exploitation and exploration, especially when dealing with a large action space; this results in the need of using pre-trained models or a mixed objective both considering rewards and classic self-supervision. In addition, the adoption of test-time metrics as reward functions might be problematic per se (see the so-called Goodhart's Law, \shortciteR{goodhart75}), while reward modeling is prone to human biases and adversarial attacks. Many challenging problems are still open, such as the integration of techniques such as IRL and multi-agent RL and the robustness of these models, in particular for preventing ``jailbreaks'' out of alignment.


\vskip 0.2in
\bibliography{biblio}
\bibliographystyle{theapa}

\end{document}